\documentclass[runningheads]{llncs}
\usepackage[T1]{fontenc}
\usepackage{graphicx,verbatim}
\usepackage{multirow} 
\usepackage[colorlinks=true,allcolors=blue]{hyperref} 
\usepackage[misc]{ifsym}
\usepackage[normalem]{ulem}
\usepackage{amsmath,amssymb,amsfonts,bm} 

\usepackage[table,xcdraw]{xcolor}
\usepackage{floatrow} 
\usepackage{booktabs} 
\usepackage{placeins}
\newfloatcommand{capbtabbox}{table}[][\FBwidth]
\begin{document}
\title{FEAT: Full-Dimensional Efficient Attention Transformer for Medical Video Generation}
%

\author{
Huihan Wang\inst{1}\textsuperscript{$\dagger$} \and
Zhiwen Yang\inst{1}\textsuperscript{$\dagger$} \and 
Hui Zhang\inst{2} \and 
Dan Zhao\inst{3} \and 
Bingzheng Wei\inst{4} \and 
Yan Xu\inst{1}\textsuperscript{(\Letter)}
} 


\authorrunning{Wang, Yang et al.}
\titlerunning{FEAT: Full-Dimensional Efficient Attention Transformer}
\institute{
School of Biological Science and Medical Engineering, State Key Laboratory of Software Development Environment, Key Laboratory of Biomechanics and Mechanobiology of Ministry of Education, Beijing Advanced Innovation Center for Biomedical Engineering, Beihang University, Beijing 100191, China
\\
\email{xuyan04@gmail.com} \and 
Department of Biomedical Engineering, Tsinghua University, Beijing 100084, China \and 
Department of Gynecology Oncology, National Cancer Center/National Clinical Research Center for Cancer/Cancer Hospital, Chinese Academy of Medical Sciences and Peking Union Medical College, Beijing 100021, China \and
ByteDance Inc., Beijing 100098, China
}

\maketitle              
\begingroup
  \renewcommand\thefootnote{}\footnote{%
    \textsuperscript{$\dagger$}   Equal contribution: H. Wang and Z. Yang. \textsuperscript{\Letter} Corresponding author: Y. Xu
  } 
\endgroup

\begin{abstract}
Synthesizing high-quality medical videos remains a significant challenge due to the need for modeling both spatial consistency and temporal dynamics. Existing Transformer-based approaches face critical limitations, including insufficient channel interactions, high computational complexity from self-attention, and coarse denoising guidance from timestep embeddings when handling varying noise levels. In this work, we propose FEAT, a full-dimensional efficient attention Transformer, which addresses these issues through three key innovations: (1) a unified paradigm with sequential spatial-temporal-channel attention mechanisms to capture global dependencies across all dimensions, (2) a linear-complexity design for attention mechanisms in each dimension, utilizing weighted key-value attention and global channel attention, and (3) a residual value guidance module that provides fine-grained pixel-level guidance to adapt to different noise levels. We evaluate FEAT on standard benchmarks and downstream tasks, demonstrating that FEAT-S, with only 23\% of the parameters of the state-of-the-art model Endora, achieves comparable or even superior performance. Furthermore, FEAT-L surpasses all comparison methods across multiple datasets, showcasing both superior effectiveness and scalability. Code is available at \href{https://github.com/Yaziwel/FEAT}{here}.



\end{abstract}
\section{Introduction}
Recent advancements in diffusion models have revolutionized artificial intelligence-generated content (AIGC) in medical imaging, enabling transformative applications in image synthesis \cite{dorjsembe2022three}, cross-modal translation \cite{wang2024soft}, and image reconstruction \cite{liu2023dolce}. While these models demonstrate remarkable capabilities in generating static medical images with spatial information, synthesizing high-fidelity dynamic medical videos—which require modeling additional temporal dynamics and consistency—remains a significant challenge. To this end, researchers have explored various approaches to encoding spatial-temporal dynamics \cite{singer2022make,khachatryan2023text2video,chen2024videocrafter2,li2024endora,xing2024make,ma2024latte,tian2025endomambaefficientfoundationmodel}, including pseudo-3D convolution \cite{singer2022make}, serial 2D+1D (spatial + temporal) convolutions \cite{xing2024make}, and spatial-temporal self-attention \cite{khachatryan2023text2video,li2024endora,chen2024videocrafter2,ma2024latte}. Given the ability of self-attention to capture long-range dependencies and the scalability of Transformers, most recent studies have largely embraced the Transformer architecture, employing cascading spatial and temporal self-attention mechanisms \cite{li2024endora,ma2024latte}.

However, the current Transformer incorporating both spatial and temporal self-attention still faces three critical limitations: \textbf{(1) Inadequate Channel-Wise Interaction.} Despite their sophisticated handling of spatial and temporal dimensions, existing architectures neglect building channel dependencies crucial for modeling feature compositions. Additionally, the impressive generation performance of diffusion models relies heavily on the denoising process while channel attention \cite{zamir2022restormer} has been widely proven to be effective for denoising. Omitting building interactions over such an important dimension hinders the model performance. \textbf{(2) Prohibitive Computational Complexity.} The self-attention mechanisms used to model both spatial and temporal dependencies suffer from quadratic computational complexity, which severely limits their practical applicability in medical videos with high resolution and many frames. \textbf{(3) Coarse Denoising Guidance.} In diffusion models, the model needs to adapt to inputs affected by different noise levels across various timesteps. Existing methods rely on timestep embeddings as global-level guidance, using adaptive layer normalization (adaLN) \cite{peebles2023scalable} to adapt to specific noise levels. However, this approach is too coarse and fails to account for dynamic interactions between noise patterns and video content. While recent work \cite{li2024endora,wang2024optical} has utilized attention maps from DINO \cite{caron2021emerging} to account for content information for finer-grained guidance, this method introduces additional substantial computational overhead during training. Therefore, existing methods have drawbacks in achieving efficient and effective medical video generation.

To address the aforementioned challenges, we present FEAT, a full-dimension efficient attention Transformer for medical video generation through three key innovations: (1) Full-Dimensional Dependency Modeling. FEAT introduces a unified paradigm with sequential spatial-temporal-channel attention, establishing global dependencies across all dimensions and enabling holistic feature modeling of medical videos. (2) Linear Complexity Design. FEAT replaces conventional self-attention with two computationally efficient components: (a) weighted key-value (WKV) attention \cite{peng2023rwkv,peng2024rwkv6,duan2024vision,yang2024restore} inspired by RWKV \cite{peng2023rwkv} for modeling spatial and temporal dependencies, and (b) global channel attention \cite{zamir2022restormer} for modeling channel dependencies. Both components achieve global dependencies within their respective dimensions while maintaining linear computational complexity \cite{shen2021linearattention}. (3) Residual Value Guidance. FEAT introduces a novel residual value guidance module (ResVGM) that leverages input embeddings—encoding both video content and specific noise patterns—as fine-grained pixel-level guidance to adapt the model for processing input of different timesteps. The ResVGM is parameter-efficient with negligible computational overhead while significantly improving generation performance. With these three innovations, FEAT achieves both efficient and effective medical video generation. Experiments show that a small version of FEAT (denoted as FEAT-S), with only 23\% of the parameters of the state-of-the-art model Endora \cite{li2024endora}, delivers comparable or even superior performance. Furthermore, the larger version, FEAT-L, outperforms all comparison methods across different datasets. 

Our contributions are three-fold:
\begin{itemize} 
\item We propose FEAT, a novel full-dimensional efficient attention Transformer for medical video generation. FEAT establishes global dependencies across all dimensions, including spatial, temporal, and channel, thereby enhancing the model’s ability to capture holistic relationships in medical videos.
\item We replace the original self-attention mechanism, which suffers from quadratic computational complexity, with attention mechanisms that establish global dependencies with linear complexity, thereby enhancing model efficiency.
\item We propose a novel residual value guidance module (ResVGM) that leverages input embeddings with both video content and specific noise patterns to provide fine-grained pixel-level guidance. This allows FEAT to effectively adapt to different timesteps with minimal computational overhead, significantly improving generation performance.
\end{itemize}

\section{Method} 
We first introduce the preliminaries of the diffusion model for video generation in Section.~\ref{sec_preliminary}. In Section~\ref{sec_feat}, we then present the details of the proposed full-dimensional efficient attention Transformer (FEAT), including its overall architecture and the specific efficient attention mechanism tailored to each dimension. Finally, in Section~\ref{sec_ResVGM}, we introduce the novel residual value guidance module (ResVGM), which provides fine-grained pixel-level guidance for adapting to different denoising timesteps.

\subsection{Preliminaries}
\label{sec_preliminary}
Diffusion probabilistic models have emerged as a groundbreaking paradigm in generative modeling, demonstrating remarkable potential for image and video synthesis. These models operate by learning to transform random noise sampled from a standard normal distribution $ p(\mathbf{x}_T) = \mathcal{N}(\mathbf{0}, \mathbf{I}) $ into high-fidelity data samples through an iterative denoising procedure. The forward diffusion process gradually corrupts input data $x_0$ by adding Gaussian noise across $T$ timesteps. This is defined by the transition $ q(\mathbf{x}_t | \mathbf{x}_{t-1}) $, with the marginal distribution at timestep $t$ expressed as: $ q(\mathbf{x}_t | \mathbf{x}_0) = \mathcal{N}(\alpha_t \mathbf{x}_0,\, \sigma_t^2 \mathbf{I}) $, where the coefficients of $ \alpha_t $ and $ \sigma_t $ are designed such that $x_T$ convergence to $ \mathcal{N}(\mathbf{0}, \mathbf{I}) $ as $ t \rightarrow T $  \cite{kingma2021variational}. In the reverse diffusion process, a noise prediction network $ \epsilon_\theta(\mathbf{x}_t, t) $ parameterizes the transition $ p(\mathbf{x}_{t-1} | \mathbf{x}_t) $, iteratively denoising $x_t$ to recover the data distribution. The training process involves optimizing the evidence lower bound (ELBO) optimization \cite{kingma2021variational}:
\begin{equation}
    \text{ELBO} = \mathbb{E}\left[\left\| \epsilon_\theta\left( \alpha_t \mathbf{x}_0 + \sigma_t \epsilon;\, t \right) - \epsilon \right\|_2^2 \right],
\end{equation}
where $ \epsilon \sim \mathcal{N}(\mathbf{0}, \mathbf{I}) $ and $ t $ follows a uniform sampling.

Since training diffusion models directly in high-resolution pixel space can be computationally expensive, we adopt the widely used approach of latent diffusion models \cite{rombach2022ldm,he2022latent}, performing the diffusion process in an encoded latent space with the help of a pretrained autoencoder \cite{rombach2022ldm} for both encoding and decoding. 

\begin{figure}[t]
\includegraphics[width=\textwidth]{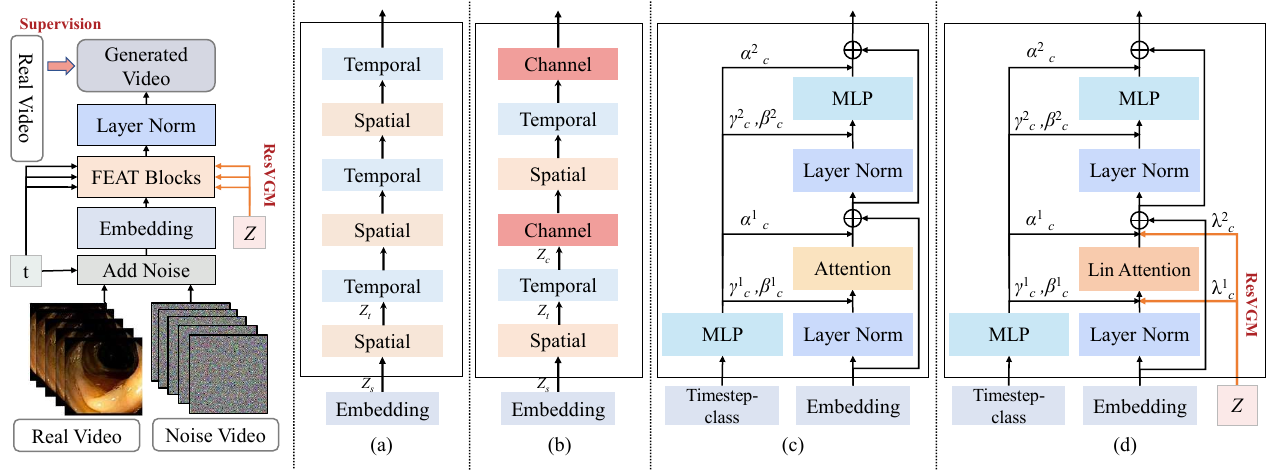}
\caption{The pipeline of FEAT for medical video generation. (a) Architecture of conventional models using cascaded spatial-temporal Transformer blocks. (b) Architecture of FEAT, which incorporates cascaded spatial-temporal-channel Transformer blocks. (c) Details of the conventional Transformer block, featuring quadratic computational complexity for self-attention and global timestep guidance. (d) Details of the Transformer block in FEAT, utilizing attention with linear computational complexity and guidance from both global timestep and pixel-level residual value $Z$.}
\label{fig1}
\end{figure}

\subsection{Full-Dimensional Efficient Attention Transformer} 
\label{sec_feat}
Existing Transformer architectures for medical video generation often face three major drawbacks: insufficient channel-wise interaction, excessive computational complexity due to self-attention, and coarse denoising guidance from the timestep. To overcome these challenges, we propose the full-dimensional efficient attention Transformer (FEAT), as illustrated in Figure.~\ref{fig1}, which introduces three key innovations: (1) Unlike conventional architectures that primarily establish spatial-temporal dependencies (as shown in Figure.~\ref{fig1} (a)), FEAT builds global dependencies across all dimensions, including spatial, temporal, and channel dimensions (as shown in Figure.~\ref{fig1} (b)). (2) To mitigate the computational burden imposed by self-attention in traditional Transformer blocks (as seen in Figure.~\ref{fig1} (c)), FEAT leverages attention mechanisms that achieve global attention with linear computational complexity across all dimensions (as shown in Figure.~\ref{fig1} (d)). (3) To address the coarse, global-level guidance that struggles to adapt to varying noise levels at different timesteps, FEAT introduces a residual value guidance module (ResVGM) for fine-grained, pixel-level denoising (as shown in Figure.~\ref{fig1} (d)). In the following subsection, we will describe the architecture of the Transformer blocks that establish spatial, temporal, and channel dependencies in detail. The details of the ResVGM are elaborated in Subsection \ref{sec_ResVGM}. 

\begin{figure}[t]
\includegraphics[width=\textwidth]{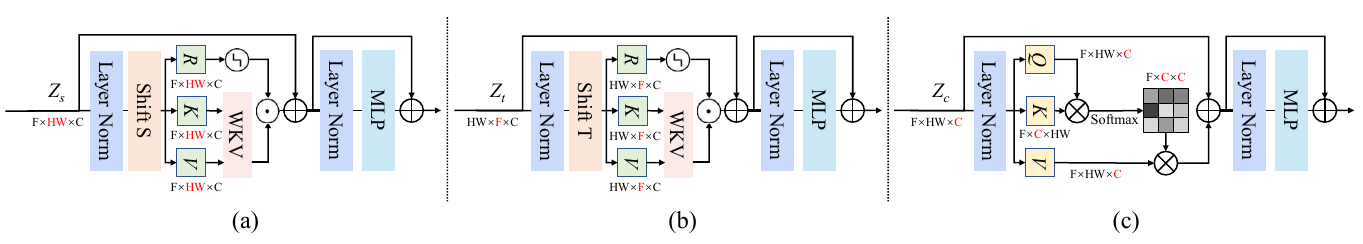}
\caption{Three distinct Transformer blocks in FEAT. (a) Spatial Transformer block with WKV attention \cite{duan2024vision}. (b) Temporal Transformer block with WKV attention \cite{duan2024vision}. (c) Channel Transformer block with global channel attention \cite{zamir2022restormer}. F, H, W, and C represent the frame number, height, width, and channel of the input feature, respectively.} 
\label{fig2}
\end{figure}

To ensure efficient modeling across all dimensions, we design different Transformer blocks with efficient attention for each dimension. Given the exceptional performance of weighted key-value (WKV) attention \cite{duan2024vision,yang2024restore} and global channel attention \cite{zamir2022restormer} in denoising, coupled with their ability to achieve global attention with linear computational complexity, we choose to apply them to denoising diffusion video generation. Specifically, \textbf{for the spatial and temporal Transformer blocks}, we adopt the WKV attention mechanism as described in \cite{duan2024vision}, as illustrated in Figure \ref{fig2} (a) and (b). To better accommodate the spatial and temporal dimensions, we modify the original token-shift mechanism from \cite{duan2024vision}, which is designed to enhance locality. For the spatial Transformer block, we introduce 2D depth-wise convolution \cite{chollet2017dwconv} (denoted 'Shift S') to strengthen locality in the spatial dimension. Similarly, for the temporal Transformer block, we apply 1D depth-wise convolution (denoted 'Shift T') to enhance locality in the temporal dimension. \textbf{For the channel Transformer block}, we directly employ the global channel attention mechanism proposed by \cite{zamir2022restormer}, as depicted in Figure \ref{fig2} (c).  With these three Transformer blocks sequentially cascaded, FEAT can efficiently establish global dependencies across spatial, temporal, and channel dimensions, enabling holistic feature modeling for medical videos.


\subsection{Residual Value Guidance Module}
\label{sec_ResVGM}
\begin{figure}[t]
\includegraphics[width=\textwidth]{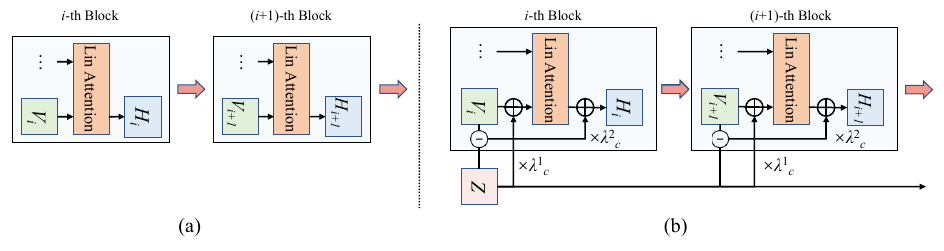}
\caption{The schematic diagram of the proposed ResVGM, where (a) represents the original framework, and (b) denotes the framework with ResVGM incorporated to different Transformer blocks. The frameworks primarily illustrate operations surrounding the attention mechanism in Transformer blocks, where ResVGM is integrated, while other modules are omitted for simplicity.} 
\label{fig3}
\end{figure}

Most existing video diffusion models employ the timestep $t$ as global guidance to adapt to specific noise levels in the denoising process.  However, this method is relatively coarse and insufficient for content-dependent denoising. To overcome this limitation, we propose integrating the input embedding as an additional, fine-grained guidance. During the denoising process, the input embedding—obtained via convolution of the input (or the denoising output at the previous timestep)—encodes both the generated video content and the associated noise patterns. These components provide crucial guidance for achieving content-dependent denoising at specific noise levels. As illustrated in Figure~\ref{fig3}, we incorporate the input embedding $Z$ to all the Transformer blocks as fine-grained guidance. Specifically, for the $i$-th Transformer block, $Z$ is added as a residual value \cite{zhou2024residual_value} to interact with the input value $V_i$ in the attention and the output hidden $H_i$ as follows:

\begin{equation}
H_i = \operatorname{LinAttention}(Q_i, K_i, V_i + \lambda^1_c Z) + \lambda^2_c (Z - V_i),
\end{equation}
where $\operatorname{LinAttention}(\cdot)$ denotes the two attention mechanisms—WKV attention and global channel attention—which both exhibit linear computational complexity. $Q_i$, $K_i$, and $V_i$ denote the query, key, and value, respectively. Note that $Q_i$ can be omitted in WKV attention. $\lambda^1_c$, $\lambda^2_c$ $\in$ $\mathbb{R}^C$ are two learnable weighting parameters. This process ensures that feature extraction across all Transformer blocks in the model is gradually refined based on the input video content and noise level. The ResVGM introduces negligible additional parameters and computational overhead, while significantly improving performance.

\section{Experiments}
\subsection{Experiment Settings}
\textbf{Datasets and Evaluation.} Our experimental evaluation is conducted on two publicly available medical video datasets: Colonoscopic \cite{mesejo2016computer} and Kvasir-Capsule \cite{borgli2020hyperkvasir}. Adhering to standardized video processing protocols \cite{ma2024latte}, we preprocess the data by uniformly extracting 16-frame sequences from continuous videos through fixed-interval sampling. All frames are resized to 128×128 pixel resolution during model training to ensure dimensional consistency. For quantitative assessment, we employ four established evaluation metrics: Fréchet Inception Distance (FID) \cite{parmar2021buggy}, Inception Score (IS) \cite{saito2017temporal}, Fréchet Video Distance (FVD) \cite{unterthiner2018towards}, and its content-debiased variant CD-FVD \cite{ge2024content}. Following the evaluation framework of StyleGAN-V \cite{skorokhodov2022stylegan}, we compute FVD scores through statistical analysis of 2048 video samples, with each sample maintaining the complete 16-frame temporal structure to preserve motion dynamics and temporal coherence.

\begin{table}[!t]
\centering
\caption{Quantitative Comparisons on Medical Video Datasets.
}\label{tab1}
\resizebox{\textwidth}{!}{

\begin{tabular}{lcccccccccccc}
\hline
                           & \multicolumn{4}{c}{Colonoscopic \cite{mesejo2016computer}}                                                                                                                                                                                         &  & \multicolumn{4}{c}{Kvasir-Capsule \cite{borgli2020hyperkvasir}}                                                                                                                                                                                      &  &                                                      &                                                      \\ \cline{2-5} \cline{7-10}
\multirow{-2}{*}{Method}   & FVD ↓                                                & CD-FVD ↓                                             & FID ↓                                                & IS ↑                                                &  & FVD ↓                                               & CD-FVD ↓                                             & FID ↓                                                & IS ↑                                                &  & \multirow{-2}{*}{Parameters(M)↓}                     & \multirow{-2}{*}{FLOPs(G)↓}                          \\ \hline
StyleGAN-V \cite{skorokhodov2022stylegan} (CVPR'22) & 2110.7                                               & 1032.8                                               & 226.14                                               & 2.12                                                &  & 183.5                                               & 898.4                                                & 31.61                                                & \cellcolor[HTML]{FFC7CE}{\color[HTML]{9C0006} 2.77} &  & \textbackslash{}                                     & \textbackslash{}                                     \\
LVDM \cite{he2022latent} (Arxiv'23)      & 1036.7                                               & 792.9                                                & 96.85                                                & 1.93                                                &  & 1027.8                                              & 615.4                                                & 200.90                                               & 1.46                                                &  & \textbackslash{}                                     & \textbackslash{}                                     \\
MoStGAN-V \cite{shen2023mostgan} (CVPR'23)  & 468.5                                                & 592.0                                                & 53.17                                                & 3.37                                                &  & 82.8                                                & 168.3                                                & 17.34                                                & 2.53                                                &  & \textbackslash{}                                     & \textbackslash{}                                     \\
Endora \cite{li2024endora} (MICCAI'24)   & \cellcolor[HTML]{C6EFCE}{\color[HTML]{006100} 460.7} & \cellcolor[HTML]{C6EFCE}{\color[HTML]{006100} 545.3} & \cellcolor[HTML]{C6EFCE}{\color[HTML]{006100} 13.41} & \cellcolor[HTML]{C6EFCE}{\color[HTML]{006100} 3.90} &  & \cellcolor[HTML]{C6EFCE}{\color[HTML]{006100} 72.3} & \cellcolor[HTML]{C6EFCE}{\color[HTML]{006100} 152.3} & \cellcolor[HTML]{C6EFCE}{\color[HTML]{006100} 10.61} & 2.54                                                &  & \cellcolor[HTML]{C6EFCE}{\color[HTML]{006100} 673.7} & \cellcolor[HTML]{FFEB9C}{\color[HTML]{9C5700} 465.8} \\ \hline
FEAT-S(Ours)               & \cellcolor[HTML]{FFEB9C}{\color[HTML]{9C5700} 415.4} & \cellcolor[HTML]{FFEB9C}{\color[HTML]{9C5700} 444.0} & \cellcolor[HTML]{FFEB9C}{\color[HTML]{9C5700} 13.34} & \cellcolor[HTML]{FFEB9C}{\color[HTML]{9C5700} 3.96} &  & \cellcolor[HTML]{FFEB9C}{\color[HTML]{9C5700} 72.2} & \cellcolor[HTML]{FFEB9C}{\color[HTML]{9C5700} 138.2} & \cellcolor[HTML]{FFEB9C}{\color[HTML]{9C5700} 9.97}  & \cellcolor[HTML]{C6EFCE}{\color[HTML]{006100} 2.65} &  & \cellcolor[HTML]{FFC7CE}{\color[HTML]{9C0006} 158.0} & \cellcolor[HTML]{FFC7CE}{\color[HTML]{9C0006} 118.7} \\
FEAT-L(Ours)               & \cellcolor[HTML]{FFC7CE}{\color[HTML]{9C0006} 351.1} & \cellcolor[HTML]{FFC7CE}{\color[HTML]{9C0006} 397.0} & \cellcolor[HTML]{FFC7CE}{\color[HTML]{9C0006} 12.31} & \cellcolor[HTML]{FFC7CE}{\color[HTML]{9C0006} 4.01} &  & \cellcolor[HTML]{FFC7CE}{\color[HTML]{9C0006} 59.2} & \cellcolor[HTML]{FFC7CE}{\color[HTML]{9C0006} 116.2} & \cellcolor[HTML]{FFC7CE}{\color[HTML]{9C0006} 8.65}  & \cellcolor[HTML]{FFEB9C}{\color[HTML]{9C5700} 2.70} &  & \cellcolor[HTML]{FFEB9C}{\color[HTML]{9C5700} 629.0} & \cellcolor[HTML]{C6EFCE}{\color[HTML]{006100} 472.1} \\ \hline
\end{tabular}

}
\end{table}

\begin{figure}[!t]
\includegraphics[width=\textwidth]{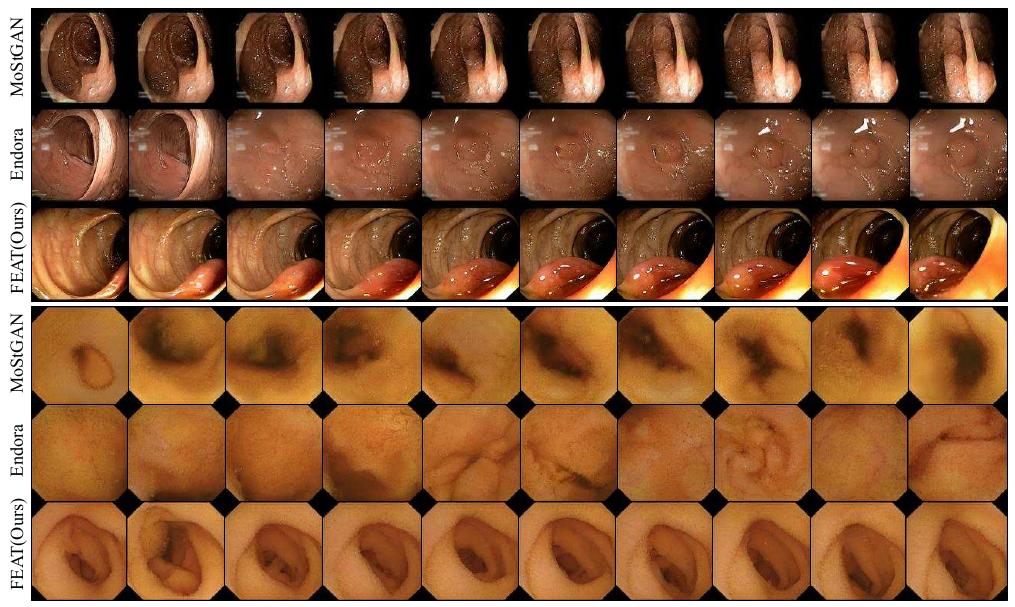}
\caption{Qualitative Comparison on Colonoscopic and Kvasir-Capsule Datasets.} \label{fig4}
\end{figure}

\noindent
\textbf{Implementation Details.} Our implementation employs the AdamW optimizer with a fixed learning rate of $1 \times 10^{-4}$ across all architectural configurations. Data preprocessing incorporates basic horizontal flipping as the sole augmentation strategy to preserve feature authenticity. The model architecture integrates a pretrained variational autoencoder from the Stable Diffusion framework \cite{blattmann2023stable} as its foundational component, enhanced by 27 specialized neural modules organized in an interleaved configuration: 9 spatial processors for geometric feature extraction, 9 temporal analyzers for motion pattern modeling, and 9 channel operators for cross-dimensional feature interaction. Hidden dimensions are configured as d=512 for small (S) models and d=1024 for large (L) model variants to accommodate computational constraints. Following established GAN training protocols \cite{peebles2023scalable}, we implement exponential moving average (EMA) stabilization \cite{li2021consistent} with all final outputs generated from converged EMA parameters, ensuring training stability and output consistency.

\subsection{Comparison with State-of-the-arts} 

We conduct performance comparison by replicating several advanced video generation models designed for general scenarios on the medical video datasets,
including StyleGAN-V \cite{skorokhodov2022stylegan}, MoStGAN-V \cite{shen2023mostgan}, LVDM \cite{he2022latent}, and Endora \cite{li2024endora}. As shown in Table.~\ref{tab1}, FEAT-S achieves comparable performance to Endora while requiring significantly fewer parameters and lower computational costs. Meanwhile, FEAT-L outperforms state-of-the-art methods. The visual qualitative comparison results in Figure.~\ref{fig4} demonstrate that FEAT can generate videos with higher quality and consistency.

\begin{table*}[t]
\begin{floatrow}
\resizebox{\textwidth}{!}{
\capbtabbox{ 
\resizebox{0.44\textwidth}{!}{
\begin{tabular}{lcl}
\hline
Method          & Colonoscopic \cite{mesejo2016computer} &  \\ \hline
Supervised-only & 74.5         &  \\
LVDM            & 76.2 (+1.7)  &  \\
Endora          & 87.0 (+12.5) &  \\ \hline
FEAT-S(Ours)    & 89.9 (+15.4) &  \\
FEAT-L(Ours)    & \textbf{91.3 (+16.8)} &  \\ \hline
\end{tabular}
}
}{
 \caption{Semi-supervised Classification Result (F1 Score) on PolyDiag \cite{tian2022contrastive} .}
 \label{tab2}
}
\capbtabbox{ 
\begin{tabular}{ccccc}
\hline
WKV                               & Channel                           & ResVGM                            & FVD↓  & FID↓  \\ \hline
{\color[HTML]{C00000} \textbf{\textit{\texttimes}}} & {\color[HTML]{C00000} \textbf{\textit{\texttimes}}} & {\color[HTML]{C00000} \textbf{\textit{\texttimes}}} & 990.0 & 23.45 \\
{\color[HTML]{70AD47} \textbf{\checkmark}} & {\color[HTML]{C00000} \textbf{\textit{\texttimes}}} & {\color[HTML]{C00000} \textbf{\textit{\texttimes}}} & 788.4 & 20.16 \\
{\color[HTML]{70AD47} \textbf{\checkmark}} & {\color[HTML]{70AD47} \textbf{\checkmark}} & {\color[HTML]{C00000} \textbf{\textit{\texttimes}}} & 583.6 & 16.98 \\ \hline
{\color[HTML]{70AD47} \textbf{\checkmark}} & {\color[HTML]{70AD47} \textbf{\checkmark}} & {\color[HTML]{70AD47} \textbf{\checkmark}} & \textbf{415.4} & \textbf{13.34} \\ \hline
\end{tabular}
}{
 \caption{Ablation Studies of Proposed Components on Colonoscopic  \cite{mesejo2016computer} Dataset.}
 \label{tab3}
 \small
}}
\end{floatrow}
\end{table*}

\subsection{Further Empirical Studies}
In this section, we demonstrate the data augmentation effects of leveraging the videos generated by our FEAT for downstream tasks, and conduct rigorous ablation experiments on the proposed improvements.\\

\noindent
\textbf{Downstream Task.} We explore the use of generated videos as unlabeled data for semi-supervised learning, specifically leveraging the FixMatch framework \cite{sohn2020fixmatch} on video-based disease diagnosis benchmarks, such as PolyDiag \cite{tian2022contrastive}. For this experiment, we randomly select 40 labeled videos ($n_l = 40$) from the PolyDiag training set and use 200 generated videos ($n_u = 200$) from Colonoscopic \cite{mesejo2016computer} as unlabeled data. The F1 scores for disease diagnosis, along with the performance improvements over the supervised-only baseline, are presented in Table.~\ref{tab2}. The results clearly demonstrate that data generated by FEAT significantly boosts the performance of downstream tasks compared to both the supervised learning baseline and other video generation techniques, thereby confirming FEAT's effectiveness as a reliable video data augmenter for video-based analysis tasks.\\

\noindent
\textbf{Ablation Studies.} Table.~\ref{tab3} presents an ablation study to evaluate the key components of the proposed FEAT-S model. We begin with a baseline that employs a simple spatial-temporal Transformer diffusion model, without incorporating any of the proposed strategies. Next, we incrementally add the three proposed design strategies: WKV attention, channel attention and ResVGM. The results clearly show that each strategy contributes to a progressive improvement in the model's performance, highlighting the essential role of these design choices in enhancing the effectiveness of the medical video generation model. 

\section{Conclusion}
This paper introduces FEAT, a novel full-dimensional efficient attention Transformer that significantly advances medical video generation. FEAT addresses three key challenges—limited channel‑wise interaction, prohibitive computational cost, and coarse denoising guidance—through three core innovations. First, a unified spatial‑temporal‑channel attention paradigm enables holistic feature modeling across all dimensions. Second, a linear‑complexity attention design makes it scale efficiently to high‑resolution videos. Third, a lightweight residual‑value guidance module adaptively refines denoising, optimizing generation performance at negligible extra computational cost. Experimental results demonstrate that FEAT outperforms existing methods in terms of both efficiency and effectiveness, marking a substantial step forward in the field of medical video generation. Future work will extend FEAT to additional imaging modalities and conduct more comprehensive evaluations \cite{ge2024content}.

\begin{credits}
\subsubsection{\ackname} 
This work is supported by the National Natural Science Foundation in China under Grant 62371016 and U23B2063, the Bejing Natural Science Foundation Haidian District Joint Fund in China under Grant L222032, the Fundamental Research Funds for the Central University of China from the State Key Laboratory of Software Development Environment in Beihang University in China, the 111 Proiect in China under Grant B13003, the SinoUnion Healthcare Inc. under the eHealth program, and the high performance computing (HPC) resources at Beihang University.

\subsubsection{\discintname}
We have no conflicts of interest to disclose.
\end{credits}
%
%
%
\bibliographystyle{splncs04}
\bibliography{ref}

\end{document}